\pdfoutput=1
\documentclass[a4paper]{article}

\usepackage{INTERSPEECH2021}
\usepackage{multirow}
\usepackage{graphicx} %use graph format
\usepackage{subfigure}
\usepackage{amsmath}

\title{Augmenting Slot Values and Contexts for Spoken Language Understanding with Pretrained Models}
\name{Haitao Lin$^{1,2}$, Lu Xiang$^{1,2}$, Yu Zhou$^{1,3}$\thanks{Yu Zhou is the corresponding author.}, Jiajun Zhang$^{1,2}$, Chengqing Zong$^{1,2}$}
\address{
  $^1$National Laboratory of Pattern Recognition, Institute of Automation, CAS, Beijing, China\\
  $^2$School of Artificial Intelligence, University of Chinese Academy of Sciences, China\\
  $^3$Fanyu AI Laboratory, Beijing Fanyu Technology Co., Ltd, Beijing, China}
\email{\{haitao.lin, lu.xiang, yzhou, jjzhang, cqzong\}@nlpr.ia.ac.cn}

\begin{document}

\maketitle
\begin{abstract}
  Spoken Language Understanding (SLU) is one essential step in building a dialogue system. Due to the expensive cost of obtaining the labeled data, SLU suffers from the data scarcity problem. Therefore, in this paper, we focus on data augmentation for slot filling task in SLU. To achieve that, we aim at generating more diverse data based on existing data. Specifically, we try to exploit the latent language knowledge from pretrained language models by finetuning them. We propose two strategies for finetuning process: value-based and context-based augmentation. Experimental results on two public SLU datasets have shown that compared with existing data augmentation methods, our proposed method can generate more diverse sentences and significantly improve the performance on SLU.
\end{abstract}

\section{Introduction}

SLU is a sub-module of dialogue system which extracts the semantic information from user inputs, including two subtasks named intent detection and slot filling. Since SLU is proved to exert significant influence on the final performance of dialogue systems~\cite{takanobu-etal-2020-goal}, improving SLU performance is a crucial problem and attracts much attention in both academia and industry. Traditionally, SLU is trained in a supervised way with sufficient labeled data, achieving excellent performance~\cite{liu2016attention,goo2018slot,wu2019joint}. Unfortunately, it is difficult and expensive to acquire enough labeled data in practice. Thus, a growing number of research focus on using few SLU data to achieve considerable performance.

A common way to boost performance with few training data is data augmentation. Data augmentation has been proved to bring about significant improvements on text classification~\cite{zhang2015character}, sentiment analysis~\cite{sun2018novel}, and spoken language understanding~\cite{kurata2016labeled,hou2018sequence}. Compared with the former two tasks, augmenting SLU data is more difficult because it needs to provide the right slot label for each word in the augmented data additionally.

In this work, we focus on data augmentation for slot filling in SLU because of its importance and difficulty under data shortage condition. According to the augmented content, we summarize data augmentation for slot filling task into two aspects: context augmentation and value augmentation. As exemplified in Table \ref{augment-example}, context augmentation focus on augmenting different sentence patterns for the same slot values. These data can increase the diversity of slot contexts and help SLU models identify slots by recognizing the contexts around them. In contrast, value augmented sentences differ from the original ones in slot values, providing different values for each slot type. SLU models can improve their ability from these new slot values.

\begin{table*}[htp]
\caption{\label{augment-example} Examples of context augmentation and value augmentation for slot filling task. Bold texts are augmented information.}
\centering
\begin{tabular}{|l|l|l|}
\hline
\multirow{2}{*}{Original data}          & Text  & Book a table somewhere in new york city for tomorrow.           \\ \cline{2-3} 
                                        & Label & city=new york city; time range=tomorrow                     \\ \hline
\multirow{2}{*}{Context augmented data} & Text  & \textbf{Please help me book a restaurant for} tomorrow \textbf{in} new york city. \\ \cline{2-3} 
                                        & Label & country=new york city; time range=tomorrow                     \\ \hline
\multirow{2}{*}{Value augmented data}   & Text  & Book a table somewhere in \textbf{san francisco} for \textbf{this evening}.            \\ \cline{2-3} 
                                        & Label & country=san francisco; time range=this evening                      \\ \hline
\end{tabular}
\end{table*}

Most of the previous work about SLU data augmentation \cite{hou2018sequence,d2019conditioned,peng2020data,Hou_Chen_Che_Chen_Liu_2021} focus on context augmentation and ignore the importance of different slot values. Although some methods \cite{kurata2016labeled,yoo2019data,shin2019utterance} try to augment slot values and contexts simultaneously in a generative method, their augmented data do not contain many new slot values in actual, since they only use the knowledge from few training data. To aim at augmenting new slot values, we try to use pretrained language models by exploiting the latent language knowledge in these models.

There exist multiple widely-used pretrained models, \textit{e.g.} BERT \cite{devlin2019bert}, GPT-2 \cite{radford2019language}, BART \cite{lewis2019bart}. Since we would like to generate a new sentence from an old one and these two sentences have much in common, it is more like a perturbation of the old sentence, similar with the training process of BART. Besides, BART has also shown its priority on augmenting other tasks \cite{kumar2020data}. Based on previous research, we propose to finetune BART model for this task. 

In this paper, we propose two different augmentation methods based on BART model: value augmentation and context augmentation, aiming at boosting diversity in two aspects. For value augmentation, we take the context information as input to generate sentences with the same contexts but different slot values. Especially, slot description is added to context information. For context augmentation, we input the slot value information and expect to obtain sentences with the same slot values but different contexts. A modified loss function is additionally proposed to help with training for both methods. Experiments on two datasets show that the value augmentation method can help improve the slot value diversity and the context augmentation method can help improve the sentence pattern diversity. Both methods achieve the most significant improvement on two SLU models compared with other augmentation methods and the mixed data of two methods can obtain better results.

\section{Related Work}

Kurata et al. \cite{kurata2016labeled} first use data augmentation method on SLU task. They try to generate diverse data by adding noises on decoder inputs, but only applying perturbation in the test phase may damage the fluency of generated sentences. Comparably, Variational AutoEncoder (VAE) can generate more various utterances by adding randomness to decoding conditions in both the train phase and the test phase. Thus, it is used in some data augmentation methods \cite{yoo2019data,shin2019utterance}. It is worth to note that all of the above methods add other parameters in the decoder to predict slot labels for generated utterances. This process may induce labeling errors, which may harm the final SLU model. Some methods \cite{hou2018sequence,d2019conditioned,peng2020data,Hou_Chen_Che_Chen_Liu_2021} instead generate delexicalized data and refill the slot values to prevent predicting slot labels. However, all of these introduced methods could not augment new slot value information which are not appeared in existing training data. Thus, we provide a new augmentation method based on pretrained models and could focus on generating new slot values as well as new contexts, filling up the blank in this area.

\section{Methods}

\subsection{Preliminary knowledge}

\subsubsection{Task definition}

In this paper, we focus on the slot filling task in SLU and its data augmentation method. For slot filling task, the aim is to extract the semantic information $s$ from a given natural language utterance $u = (w_1, ..., w_n)$. The semantic information is expressed in slot type and value pairs $\{(t_1, v_1), ..., (t_k, v_k)\}$. 
%Since all the slot values appear in utterance $u$, we can treat slot filling task as a sequence labeling task where we output each word $w_i$ with a label in a BIO scheme.
Given the train set $D = \{(u^{(1)}, s^{(1)}), ..., (u^{(m)}, s^{(m)})\}$, data augmentation enlarges the training set $D$ by adding new labeled data, which is then used to train an SLU model. 
%we expect the model can achieve higher performance by using the augmented data.

\subsubsection{BART pretrained model}

BART is a denoising sequence-to-sequence pretraining model used for natural language understanding and generation. It is composed of an encoder and decoder, both based on the transformer structure. 
The model accepts a natural language sentence as input and generates a sentence as output. During the pretraining period, the input texts are corrupted with some noising process and the model is trained to reconstruct the original texts. This pretraining strategy makes the model obtain the ability of language understanding and generation. More details of BART can be found in the original paper \cite{lewis2019bart}.

\subsection{Model description}

In this section, we will introduce two different data augmentation strategies based on BART model. Each strategy is introduced with its input transformation operation, augmentation procedure, and data filtering method.

\subsubsection{Value-based augmentation}
\begin{figure*}[htbp] 
    \centering  
    \includegraphics[scale=0.45]{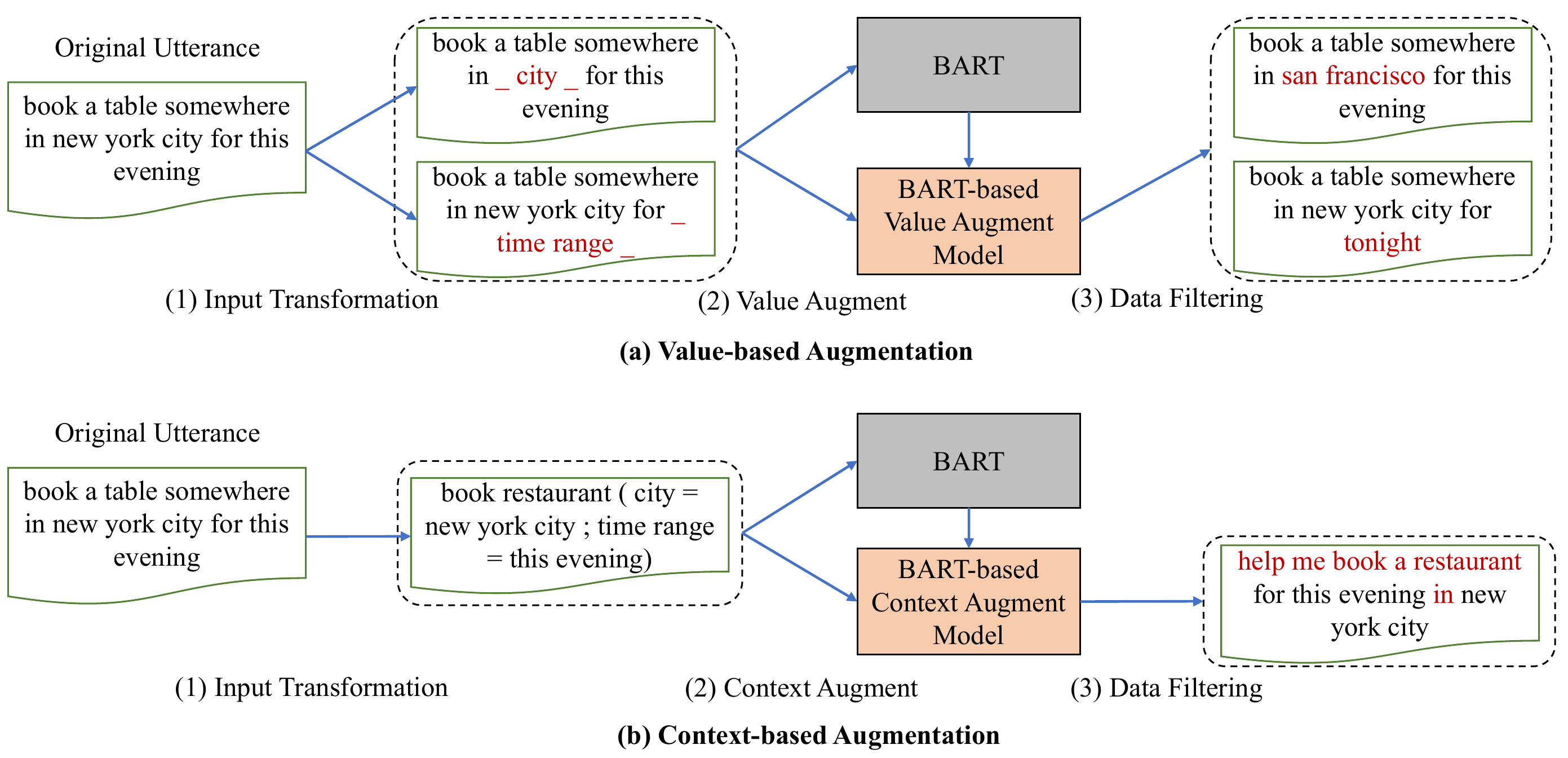} 
    \caption{Our proposed data augmentation methods. The upper part is value augmentation and the lower part is context augmentation.} 
    \label{fig:model}
\end{figure*}

We depict our value-based augmentation model structure in Figure \ref{fig:model} (a). The first step of our method is to transform the training data. Given an utterance $u$ and its slot information $s = \{(t_1, v_1), ..., (t_k, v_k)\}$ with k different slots, we randomly choose a slot type $t_j$ and mask the value $v_j$ appeared in $u$. Instead of using a regular ``\textit{[MASK]}'' token, we replace $v_j$ with the natural language description of $t_j$. We find that the mask token may let the model generate inappropriate slot values that belong to other slot types, since it loses the information of the original slot type. In comparison, slot description can make the model understand the semantic information of the chosen slot and generate the slot value more correctly. Additionally, we add a special token ``$\_$'' at the start and the end of $t_j$ to let the model know the position of the replaced slot type. After that, we take the modified sentence as model input and the original utterance $u$ as output. As shown in Figure \ref{fig:model} (a), if we choose the slot ``\textit{city}'', the modified input becomes ``\textit{book a table somewhere in $\_$ city $\_$ \ for this evening}''; If we choose the slot ``\textit{time$\_$range}'', the input becomes ``\textit{book a table somewhere in new york city  for $\_$ time range $\_$}''. For an utterance with k slots, we can generate k different inputs for training.

After obtaining the modified data for training, the model is trained by optimizing a reconstruction loss between the decoder's output and the ground truth, which is the original utterance. Specially, we use a \textbf{Modified Label Smoothing Cross Entropy} loss to boost the diversity of generated slot values and guarantee the output quality. As presented in Equation 1 and 2, it is a variant of cross entropy loss with different ground truth labels. 
$w_i$ is the $i^{th}$ word in $u$ and $v_j$ is the delexicalized slot value. The predicted distribution for $w_i$ is $\hat{y_i}$. $\epsilon$ is a parameter for label smoothing and $|V|$ is vocabulary size. The key point of the proposed method is to only use label smoothing strategy \cite{szegedy2016rethinking} on slot value words, forcing the model to generate more diverse slot values but not other contexts. At the test time, we use the same modified input as in the training period and autogressively decode the predicted sentence. We wish that the predicted sentence has the same context with original utterance $u$ but differs in the chosen slot value $v_j$.

\begin{align}
\textit{Loss} &= -\sum_{i=1}^n{y'_i \cdot \log(\hat{y_i})} \\
y'_i &= \left\{
\begin{aligned}
&[0, \cdots, 1, \cdots, 0] & w_i \notin v_j \\
&[\frac{\epsilon}{|V|-1}, \cdots, 1-\epsilon, \cdots, \frac{\epsilon}{|V|-1}] & w_i \in v_j
\end{aligned}
\right.
\end{align}

The last process is to filter out low-quality sentences from the generated data and assign high-quality ones with correct labels. We compare the generated output with its delexicalized input and filter out the data that do not have the same contexts. Besides, we can also recognize the generated slot values by a simple string comparison algorithm, as marked with red color in Figure \ref{fig:model} (a). Then we assign context words in the generated sentence with their original labels and the slot values with the chosen slot type label $t_j$. Since our method only delexicalizes one slot and predict one new slot value, it makes the labeling step much easier and the result more reliable.

\subsubsection{Context-based augmentation}

Figure \ref{fig:model} (b) presents the structure of our context-based augmentation model. Similar with the natural language generation task, we send the intent and slot information to the encoder and let the decoder output the natural language sentence. For utterance $u$ with intent $I$, the slot information of $u$ is formatted as $I (t_1 = v_1 ; ... ; t_n = v_n)$. Thus, the input of the given example becomes ``\textit{book restaurant (city = new york city ; time range = this evening)}''. The output of the model for training is also the original utterance $u$.

All the training settings are consistent with those in value-based augmentation. There exists one difference that we use the label smoothing strategy on context words instead of slot value words since we wish the decoder to generate sentences with the same slot values as the original data but different contexts.

As for data filtering, we filter out generated sentences that do not contain all the slot values existed in the input through a matching process. Besides, we also filter out sentences that contain slot values more than needed by comparing generated sentences with a slot dictionary acquired from the given training data. The same slot labels of the input can be assigned to the generated data after the above filtering process.

\section{Experiments}
\label{sec:exp}

\subsection{Datasets}

In this work, we employ two widely used SLU datasets, including ATIS \cite{hemphill1990atis} and Snips \cite{coucke2018snips}, to verify the data augmentation performance. ATIS only contains flight reservation related requests, while Snips contains more domains such as music, movie, and restaurant.

In order to test the data augmentation performance with few training data, we split the dataset into small dataset (1/40 of the full data) and medium dataset (1/10 of the full data) to test different methods, similar with the previous work \cite{hou2018sequence}. Since there is no publicly available split for experiment, we split the dataset randomly and test all the methods on our splits. Besides, all of our data\footnote{https://github.com/MiuLab/SlotGated-SLU} are in lower-case, which is more usual in SLU settings. Our data splits and codes are publicly available\footnote{https://github.com/xiaolinAndy/SLU-Aug-PrLM}. 

\subsection{Baselines and experiment settings}
\label{subsec:baselines}

We choose some previous methods as baselines for comparison. The baselines include \textbf{No Augmentation}, \textbf{Seq2seq Augmentation} \cite{hou2018sequence}, \textbf{VAE} \cite{d2019conditioned}, and \textbf{GPT-2} \cite{peng2020data}. Our proposed methods are noted as \textbf{BART Value} and \textbf{BART Context}, representing augmenting slot values and contexts respectively. 

For all baseline methods, we augment the same amount of data as the given dataset, thus doubling the training data. We validate the performance of Seq2seq Augmentation %\footnote{https://github.com/AtmaHou/Seq2SeqDataAugmentationForLU}
and VAE %\footnote{https://github.com/snipsco/automatic-data-generation}
methods using publicly available codes. GPT-2 method is re-implemented according to the description in the paper \cite{peng2020data}. 

For BART-based methods, we use bart-large\footnote{https://github.com/huggingface/transformers}
pretrained model and default finetuning parameters. After achieving the best perplexity on validation set, we use the finetuned model to generate sentences. The label smoothing parameter is 0.1.

% Besides, we also experiment on extra training data to test the theoretical performance of augmentation effectiveness in two aspects (value and context). The experiment includes three following settings:

% \textbf{Value Augment Extra}: We treat the full training dataset as extra data. In this setting, we augment the given training data by replacing the slot values with the possible alternatives in the full training dataset. Since full training dataset guarantees the credibility and diversity of augmented data, it can be approximately treated as the upper bound of value-based augmentation methods.

% \textbf{Context Augment Extra}: We randomly choose some of the delexicalized utterances from full training data. Then we replace these delexicalized slots with appropriate values appeared in the given training data. It can be treated as the upper bound of context-based augmentation methods.

% \textbf{Full Augment Extra}: We simply use some labeled data from the full dataset as the augmented data, thus augmenting both context and value simultaneously. It can be treated as the upper bound of all augmentation methods. 

\begin{table*}[htbp]
\centering
\small
\caption{\label{given-data-result} The SLU performance of different data augmentation methods using the given training dataset.}
\begin{tabular}{cllllllll}
\hline
Dataset         & \multicolumn{4}{c}{ATIS}                                         & \multicolumn{4}{c}{Snips}                                         \\ \hline
Split           & \multicolumn{2}{c}{Small (111)} & \multicolumn{2}{c}{Medium (447)} & \multicolumn{2}{c}{Small (327)} & \multicolumn{2}{c}{Medium (1308)} \\ \hline
SLU model       & LSTM           & BERT          & LSTM           & BERT           & LSTM           & BERT          & LSTM            & BERT           \\ \hline
No augmentation & 71.26          & 82.50         & 86.39          & 90.72          & 58.61          & 74.36         & 77.14           & 89.82          \\
Seq2Seq         & 71.53          & 83.44         & 87.37          & 91.34          & 58.66          & 74.75         & 78.56           & 91.00          \\
VAE             & 71.77               & 83.32              & \textbf{88.28}                & 91.31                & 59.64                & 75.73              &  78.96               & 90.12              \\
GPT-2        & 71.68      & 82.97    & 87.44      & 91.41      & 59.37       & 74.55     & 78.59       & 90.74      \\
BART Value      & 73.06         & \textbf{83.60}         & 87.31          & 91.61          & \textbf{61.06}          & \textbf{77.23}         & \textbf{79.06}           & \textbf{91.13}          \\
BART Context    & \textbf{73.95}               &  83.43             & 87.89                 & \textbf{91.94}               & 60.73               & 75.60              & 79.03                & 90.34               \\ \hline
\end{tabular}
\end{table*}

\begin{table}[htp]
\centering
\small
\caption{\label{full-data-result} The SLU performance of different data augmentation methods on full Snips data.}
\begin{tabular}{cll}
\hline
Dataset         & \multicolumn{2}{c}{Snips} \\ \hline
SLU model       & LSTM        & BERT        \\ \hline
No augmentation & 91.69       & \textbf{96.51}       \\
Seq2Seq         & 92.28       & 96.31       \\
VAE             & 91.15       & 96.08            \\
GPT-2        &    91.74    & 96.45 \\
BART Value      & \textbf{92.48}       & 96.42       \\
BART Context    & 91.83       & 96.20            \\ \hline
\end{tabular}
\end{table}

\subsection{Evaluation}
\label{sec:eval}

We use two classic and widely-used SLU models to test the efficiency of different methods, making the result more convincing. One is a single layer Bi-LSTM model%\footnote{https://github.com/AtmaHou/Bi-LSTM$\_$PosTagger}
, adopted by previous work \cite{hou2018sequence}. The other one is a finetuned BERT model for slot filling \cite{chen2019bert}. Compared with LSTM model, BERT model could achieve higher performance with few training data, which makes data augmentation more challenging. After training each model with the augmented data, we calculate the entity-level F1 score of slot filling. We run each SLU model for five times and take the average result to eliminate randomness.
 
 %\footnote{https://github.com/sz128/slot$\_$filling$\_$and$\_$intent$\_$detection$\_$of$\_$SLU}

In addition to SLU performance, we also want to evaluate the diversity in augmented data. Two metrics are adopted here: (1) \textbf{Word diversity} measures text diversity by counting the proportion of different words in the augmented data compared with the original data. (2) \textbf{Originality} represents the proportion of new data compared with the given dataset. It is calculated on delexicalized sentences to show the diversity of contexts.

\subsection{Analysis}
 
% To analyze the effect of different data augmentation methods, we focus on four main questions: (1) Which method performs the best among different data augmentation methods? (2) Do different datasets or different SLU models influence data augmentation performance? (3) Will the data augmentation methods do harm to SLU model performance when we have enough training data? (4) Do our proposed preprocessing method and loss function actually work?

\subsubsection{Augmentation performance}
We summarize all the results in Table \ref{given-data-result}. First, the BART Value method achieves the best performance on all settings of Snips and the BART Context method performs the best on two settings of ATIS. Both two proposed methods perform better than traditional methods (Seq2seq and VAE) and methods using extra knowledge (GPT-2) on most of the conditions. We infer that the text infilling pretraining process of BART model leads to better performance since it is similar with our finetuning method. In all, the result shows the effectiveness of our proposed methods.

From the results of two SLU models, we can conclude that LSTM-based model benefits more than BERT-based model from data augmentation methods, mainly because LSTM-based model has a lower baseline performance. For different datasets, Snips benefits from data augmentation more than ATIS. Specifically, BART Value method works well on Snips, while BART Context method performs better on ATIS. We believe that it is due to the vocabulary difference. Snips has a larger vocabulary and slot values are thus more critical on predicting slot labels. Conversely, ATIS has a smaller vocabulary. Therefore, slot values are less influential than slot contexts.

To figure out whether these methods still work when enough training data is given, we run them on the full Snips training data and show the result in Table \ref{full-data-result}. Almost all the data augmentation methods contribute some improvements on LSTM model and slightly degradation on BERT model, mainly due to the robustness of BERT model. Still, our proposed BART Value method performs the best among all methods in general.

In Table \ref{tab:ablation}, we do some ablation test to verify the effectiveness of the two settings in BART Value method. First, we remove the modified label smoothing method and smooth all the labels. Second, we use token ``\textit{[MASK]}'' to represent the slot type instead of the slot description. Both variants perform worse than the original one, proving that both the two settings help our model generate more useful data for SLU models. Besides, we also mix the augmented data of BART Value method and BART Context method, and test the SLU performance on the mixed data. As expected, the mixed data achieves higher performance than each of the single data, showing that two methods help with SLU models in different ways and it is better to augment both aspects for achieving the best SLU performance.

\begin{table}[htp]
\centering
\small
\caption{\label{tab:ablation} Performance of some variants of BART Value method and the mixed data.}
\begin{tabular}{lll}
\hline
Dataset                    & \multicolumn{2}{c}{Snips (medium)} \\ \hline
SLU model                  & LSTM            & BERT            \\ \hline
BART Value                 & 79.06           & 91.13           \\
- modified label  & 78.50 (-0.56)    & 91.08 (-0.05)    \\
- slot description         & 74.27 (-4.79)    & 87.88 (-3.25)    \\ \hline
BART Context               & 79.03           & 90.34           \\
Mixed data                 & 79.39           & 91.13              \\ \hline
\end{tabular}
\end{table}

\subsubsection{Diversity Analysis}
In this section, we compare the diversity of augmented data according to the two metrics mentioned in Section \ref{sec:eval}. As shown in Table \ref{analysis-data}, our proposed methods can generate more diverse data compared with the given training data. BART Value method achieves the highest word diversity by augmenting more slot values that do not appear in the original dataset.
Additionally, BART Context method achieves the highest originality, representing the highest variety of contexts in the augmented dataset.

%we calculate the word diversity and the originality of delexicalized sentences. The result proves that

\begin{table}[htp]
\centering
\small
\caption{\label{analysis-data} The diversity results of different augmentation methods. Note that since BART Value method only augment the slot value, its originality of delexicalized sentences is 0.}
\begin{tabular}{ccc}
\hline
Dataset      & \multicolumn{2}{c}{Snips (medium)}                                         \\ \hline
Metrics      & word diversity & originality-delex \\ \hline
Seq2Seq      & 0.13               & 8.72 \\
VAE          & 0.13               & 49.46 \\
BART Value   &  \textbf{2.67}      & 0 \\
BART Context &  0.02     & \textbf{65.52} \\ \hline
\end{tabular}
\end{table}

\section{Conclusions}

In this paper, we propose two data augmentation strategies for SLU by augmenting slot values and contexts based on BART pretrained model. Experimental results on ATIS and Snips datasets prove that our proposed model can improve the SLU model performance to a larger extent than other augmentation methods and the generated sentences also have higher diversity. In the future, we plan to further our work by merging value augmentation and context augmentation into a single model and trying out other pretrained models for SLU data augmentation.

\section{Acknowledgements}

We thank anonymous reviewers for helpful suggestions. The research work described in this paper has been supported by the National Key Research and Development Program of China under Grant No. 2017YFB1002103.

\bibliographystyle{IEEEtran}

\bibliography{mybib}

\begin{thebibliography}{10}
\providecommand{\url}[1]{#1}
\csname url@samestyle\endcsname
\providecommand{\newblock}{\relax}
\providecommand{\bibinfo}[2]{#2}
\providecommand{\BIBentrySTDinterwordspacing}{\spaceskip=0pt\relax}
\providecommand{\BIBentryALTinterwordstretchfactor}{4}
\providecommand{\BIBentryALTinterwordspacing}{\spaceskip=\fontdimen2\font plus
\BIBentryALTinterwordstretchfactor\fontdimen3\font minus
  \fontdimen4\font\relax}
\providecommand{\BIBforeignlanguage}[2]{{%
\expandafter\ifx\csname l@#1\endcsname\relax
\typeout{** WARNING: IEEEtran.bst: No hyphenation pattern has been}%
\typeout{** loaded for the language `#1'. Using the pattern for}%
\typeout{** the default language instead.}%
\else
\language=\csname l@#1\endcsname
\fi
#2}}
\providecommand{\BIBdecl}{\relax}
\BIBdecl

\bibitem{takanobu-etal-2020-goal}
R.~Takanobu, Q.~Zhu, J.~Li, B.~Peng, J.~Gao, and M.~Huang, ``Is your
  goal-oriented dialog model performing really well? empirical analysis of
  system-wise evaluation,'' in \emph{Proceedings of the 21th Annual Meeting of
  the Special Interest Group on Discourse and Dialogue}.\hskip 1em plus 0.5em
  minus 0.4em\relax 1st virtual meeting: Association for Computational
  Linguistics, Jul. 2020, pp. 297--310.

\bibitem{liu2016attention}
B.~Liu and I.~Lane, ``Attention-based recurrent neural network models for joint
  intent detection and slot filling,'' \emph{Interspeech 2016}, pp. 685--689,
  2016.

\bibitem{goo2018slot}
C.-W. Goo, G.~Gao, Y.-K. Hsu, C.-L. Huo, T.-C. Chen, K.-W. Hsu, and Y.-N. Chen,
  ``Slot-gated modeling for joint slot filling and intent prediction,'' in
  \emph{Proceedings of the 2018 Conference of the North {A}merican Chapter of
  the Association for Computational Linguistics: Human Language Technologies,
  Volume 2 (Short Papers)}.\hskip 1em plus 0.5em minus 0.4em\relax New Orleans,
  Louisiana: Association for Computational Linguistics, Jun. 2018, pp.
  753--757.

\bibitem{wu2019joint}
J.~Wu, L.~F. D'Haro, N.~F. Chen, P.~Krishnaswamy, and R.~E. Banchs, ``Joint
  learning of word and label embeddings for sequence labelling in spoken
  language understanding,'' in \emph{2019 IEEE Automatic Speech Recognition and
  Understanding Workshop (ASRU)}.\hskip 1em plus 0.5em minus 0.4em\relax IEEE,
  2019, pp. 800--806.

\bibitem{zhang2015character}
X.~Zhang, J.~J. Zhao, and Y.~LeCun, ``Character-level convolutional networks
  for text classification,'' in \emph{Advances in Neural Information Processing
  Systems 28: Annual Conference on Neural Information Processing Systems 2015,
  December 7-12, 2015, Montreal, Quebec, Canada}, C.~Cortes, N.~D. Lawrence,
  D.~D. Lee, M.~Sugiyama, and R.~Garnett, Eds., 2015, pp. 649--657.

\bibitem{sun2018novel}
X.~Sun and J.~He, ``A novel approach to generate a large scale of supervised
  data for short text sentiment analysis,'' \emph{Multimedia Tools and
  Applications}, pp. 1--21, 2018.

\bibitem{kurata2016labeled}
G.~Kurata, B.~Xiang, and B.~Zhou, ``Labeled data generation with
  encoder-decoder lstm for semantic slot filling.'' in \emph{INTERSPEECH},
  2016, pp. 725--729.

\bibitem{hou2018sequence}
Y.~Hou, Y.~Liu, W.~Che, and T.~Liu, ``Sequence-to-sequence data augmentation
  for dialogue language understanding,'' in \emph{Proceedings of the 27th
  International Conference on Computational Linguistics}.\hskip 1em plus 0.5em
  minus 0.4em\relax Santa Fe, New Mexico, USA: Association for Computational
  Linguistics, Aug. 2018, pp. 1234--1245.

\bibitem{d2019conditioned}
S.~d'Ascoli, A.~Coucke, F.~Caltagirone, A.~Caulier, and M.~Lelarge,
  ``Conditioned query generation for task-oriented dialogue systems,''
  \emph{arXiv preprint arXiv:1911.03698}, 2019.

\bibitem{peng2020data}
B.~Peng, C.~Zhu, M.~Zeng, and J.~Gao, ``Data augmentation for spoken language
  understanding via pretrained models,'' \emph{arXiv preprint
  arXiv:2004.13952}, 2020.

\bibitem{Hou_Chen_Che_Chen_Liu_2021}
Y.~Hou, S.~Chen, W.~Che, C.~Chen, and T.~Liu, ``C2c-genda: Cluster-to-cluster
  generation for data augmentation of slot filling,'' \emph{Proceedings of the
  AAAI Conference on Artificial Intelligence}, vol.~35, no.~14, pp.
  13\,027--13\,035, May 2021.

\bibitem{yoo2019data}
K.~M. Yoo, Y.~Shin, and S.-g. Lee, ``Data augmentation for spoken language
  understanding via joint variational generation,'' in \emph{Proceedings of the
  AAAI Conference on Artificial Intelligence}, vol.~33, 2019, pp. 7402--7409.

\bibitem{shin2019utterance}
Y.~Shin, K.~M. Yoo, and S.-G. Lee, ``Utterance generation with variational
  auto-encoder for slot filling in spoken language understanding,'' \emph{IEEE
  Signal Processing Letters}, vol.~26, no.~3, pp. 505--509, 2019.

\bibitem{devlin2019bert}
J.~Devlin, M.-W. Chang, K.~Lee, and K.~Toutanova, ``{BERT}: Pre-training of
  deep bidirectional transformers for language understanding,'' in
  \emph{Proceedings of the 2019 Conference of the North {A}merican Chapter of
  the Association for Computational Linguistics: Human Language Technologies,
  Volume 1 (Long and Short Papers)}.\hskip 1em plus 0.5em minus 0.4em\relax
  Minneapolis, Minnesota: Association for Computational Linguistics, Jun. 2019,
  pp. 4171--4186.

\bibitem{radford2019language}
A.~Radford, J.~Wu, R.~Child, D.~Luan, D.~Amodei, and I.~Sutskever, ``Language
  models are unsupervised multitask learners,'' \emph{OpenAI blog}, vol.~1,
  no.~8, p.~9, 2019.

\bibitem{lewis2019bart}
M.~Lewis, Y.~Liu, N.~Goyal, M.~Ghazvininejad, A.~Mohamed, O.~Levy, V.~Stoyanov,
  and L.~Zettlemoyer, ``{BART}: Denoising sequence-to-sequence pre-training for
  natural language generation, translation, and comprehension,'' in
  \emph{Proceedings of the 58th Annual Meeting of the Association for
  Computational Linguistics}.\hskip 1em plus 0.5em minus 0.4em\relax Online:
  Association for Computational Linguistics, Jul. 2020, pp. 7871--7880.

\bibitem{kumar2020data}
V.~Kumar, A.~Choudhary, and E.~Cho, ``Data augmentation using pre-trained
  transformer models,'' in \emph{Proceedings of the 2nd Workshop on Life-long
  Learning for Spoken Language Systems}.\hskip 1em plus 0.5em minus 0.4em\relax
  Suzhou, China: Association for Computational Linguistics, Dec. 2020, pp.
  18--26.

\bibitem{szegedy2016rethinking}
C.~Szegedy, V.~Vanhoucke, S.~Ioffe, J.~Shlens, and Z.~Wojna, ``Rethinking the
  inception architecture for computer vision,'' in \emph{2016 {IEEE} Conference
  on Computer Vision and Pattern Recognition, {CVPR} 2016, Las Vegas, NV, USA,
  June 27-30, 2016}.\hskip 1em plus 0.5em minus 0.4em\relax {IEEE} Computer
  Society, 2016, pp. 2818--2826.

\bibitem{hemphill1990atis}
C.~T. Hemphill, J.~J. Godfrey, and G.~R. Doddington, ``The {ATIS} spoken
  language systems pilot corpus,'' in \emph{Speech and Natural Language:
  Proceedings of a Workshop Held at Hidden Valley, {P}ennsylvania, June
  24-27,1990}, 1990.

\bibitem{coucke2018snips}
A.~Coucke, A.~Saade, A.~Ball, T.~Bluche, A.~Caulier, D.~Leroy, C.~Doumouro,
  T.~Gisselbrecht, F.~Caltagirone, T.~Lavril \emph{et~al.}, ``Snips voice
  platform: an embedded spoken language understanding system for
  private-by-design voice interfaces,'' \emph{arXiv preprint arXiv:1805.10190},
  2018.

\bibitem{chen2019bert}
Q.~Chen, Z.~Zhuo, and W.~Wang, ``Bert for joint intent classification and slot
  filling,'' \emph{arXiv preprint arXiv:1902.10909}, 2019.

\end{thebibliography}

% \begin{thebibliography}{9}
% \bibitem[1]{Davis80-COP}
%   S.\ B.\ Davis and P.\ Mermelstein,
%   ``Comparison of parametric representation for monosyllabic word recognition in continuously spoken sentences,''
%   \textit{IEEE Transactions on Acoustics, Speech and Signal Processing}, vol.~28, no.~4, pp.~357--366, 1980.
% \bibitem[2]{Rabiner89-ATO}
%   L.\ R.\ Rabiner,
%   ``A tutorial on hidden Markov models and selected applications in speech recognition,''
%   \textit{Proceedings of the IEEE}, vol.~77, no.~2, pp.~257-286, 1989.
% \bibitem[3]{Hastie09-TEO}
%   T.\ Hastie, R.\ Tibshirani, and J.\ Friedman,
%   \textit{The Elements of Statistical Learning -- Data Mining, Inference, and Prediction}.
%   New York: Springer, 2009.
% \bibitem[4]{YourName17-XXX}
%   F.\ Lastname1, F.\ Lastname2, and F.\ Lastname3,
%   ``Title of your INTERSPEECH 2021 publication,''
%   in \textit{Interspeech 2021 -- 20\textsuperscript{th} Annual Conference of the International Speech Communication Association, September 15-19, Graz, Austria, Proceedings, Proceedings}, 2020, pp.~100--104.
% \end{thebibliography}

\end{document}